\documentclass{IEEE_lsens}
\usepackage{textcomp}

\usepackage[noadjust]{cite}

\ifCLASSINFOpdf
\else
\fi
\usepackage[T1]{fontenc} 
\usepackage{amsmath}
\usepackage[cmintegrals]{newtxmath}
\usepackage{bm}

\usepackage{array}
\usepackage{url}
\ifCLASSINFOpdf
\else
\fi
\providecommand{\hypersetup}[1]{\relax}

\hypersetup{pdftitle={Bare Demo of IEEE\_lsens.cls for IEEE Sensors Letters},
pdfsubject={Typesetting},
pdfauthor={Michael D. Shell},
pdfkeywords={Class, IEEE, IEEE\_lsens, IEEE Sensors Letters, LaTeX, Typesetting, TeX}}
\usepackage[utf8]{inputenc}
\usepackage{graphicx}

\begin{document}

\markboth{Vol.~x, No.~xx, xxx~2022}{0000000}

\IEEELSENSarticlesubject{Sensor Applications}

%
\title{Learning Car Speed Using Inertial Sensors for Dead Reckoning Navigation}

%
\author{\IEEEauthorblockN{Maxim Freydin, Member, IEEE and Barak Or, Member, IEEE
}
\IEEEauthorblockA{Maxim Freydin and Barak Or are with ALMA Technologies, Haifa, 3100400, Israel\\
}%
\thanks{Corresponding author: M. Freydin (e-mail: maxim@almatechnologies.com).}
\thanks{Associate Editor: XXXX XXXX.}%
\thanks{Digital Object Identifier 10.1109/LSENS.2017.0000000}}
%
%
%

\IEEELSENSmanuscriptreceived{Manuscript accepted to IEEE Sensors Letters at August 22, 2022; This is a preprint version}

\IEEEtitleabstractindextext{%
\begin{abstract}
A deep neural network (DNN) is trained to estimate the speed of a car driving in an urban area using as input a stream of measurements from a low-cost six-axis inertial measurement unit (IMU). Three hours of data was collected by driving through the city of Ashdod, Israel in a car equipped with a global navigation satellite system (GNSS) real time kinematic (RTK) positioning device and a synchronized IMU. Ground truth labels for the car speed were calculated using the position measurements obtained at the high rate of 50 Hz. A DNN architecture with long short-term memory layers is proposed to enable high-frequency speed estimation that accounts for previous inputs history and the nonlinear relation between speed, acceleration and angular velocity. A simplified aided dead reckoning localization scheme is formulated to assess the trained model which provides the speed pseudo-measurement. The trained model is shown to substantially improve the position accuracy during a 4 minutes drive without the use of GNSS position updates.\end{abstract}

}


\maketitle

\section{Introduction}
Accurate localization is a critical capability for automated driving \cite{luettel2012autonomous,badue2021self,rana2017attack,chan2020framework}. Existing solutions are based on sensor fusion techniques that utilize inputs from cameras, radars, global navigation satellite system (GNSS) receivers, and inertial measurement units (IMUs) \cite{gustafsson2002particle,mikov2021stereo,farrell2008aided}. However, some environments, like indoor parking lots and dense urban areas, remain challenging to localize in. Cameras and radars require line-of-sight toward the object of interest and GNSS receivers can not operate indoor. IMU sensors, on the other hand, are immune to those problems and remain available in all settings. The shortcoming of IMUs for localization is that existing integration methods accumulate error rapidly over time \cite{liu2021bias,guerrier2016theoretical}.

In this work, a deep neural network (DNN)  \cite{lecun2015deep} model is trained to predict the speed of a car from IMU measurements which may then be used for localization (by an additional integration) or used as speedometer. This work is aimed at creating a high-frequency real-time speed estimator using a single low-cost IMU sensor. For that, we establish a data-driven approach that consists of two steps: (1) collecting data of IMU signals from multiple car drives with corresponding speed measurements as labels, using a real time kinematic (RTK)-GNSS device, and (2) applying supervised learning to train a DNN-based model to estimate the car speed in real-time. The proposed approach may improve existing dead reckoning (DR) techniques in GNSS-denied areas \cite{luo2018geographical} and enable localization based only on a single IMU sensor for longer time periods while maintaining an acceptable positioning error.

Previous attempts to apply supervised learning in estimation problems with localization applications were focused on learning the sensor noise models. Mahdi et al. used low-cost sensors as input together with expensive sensors to generate the true labels for the measurement noise \cite{mahdi2022}. Azzam et al. trained long short-term memory (LSTM) models to improve the trajectory accuracy obtained using a simultaneous localization and mapping (SLAM) algorithm \cite{azzam2021}. However, they used the pose estimated by the SLAM algorithm as input and not direct sensor measurements. Srinivasan et al. \cite{srinivasan2020} learned speed estimation for an autonomous racing car. The true label for the speed was produced using a specially developed mixed Kalman Filter (KF) \cite{welch1995introduction} to fuse inputs from multiple sensors of different types. In \cite{Reina2018DEEPLB}, Reina et al. used a convolutional neural network (CNN)-based architecture to train a regressor to predict speed. However, this approach is not able to create memory between samples. Also, the large window size of the input makes the model unfit for high-frequency real-time applications. Yanlei et al. used a similar approach to estimate speed with low-cost IMU as input \cite{yanlei2019}. Their solution was based on position measurements from a common GNSS receiver to calculate the true speed which suffers from relatively low accuracy and low sampling rate. In addition, the input was the IMU signal sampled at the low frequency of 10 Hz and a window size of $4$ seconds which is not sufficient for most real-time applications. Karlsson et al. also used a CNN-based architecture to train a regressor to predict speed from an accelerometer and the yaw rate \cite{karlsson2021speed}. They also considered a large input time window (more than 2 seconds) and an architecture which is not able to establish memory between samples. This work aims to address the shortcomings of previous work and apply the trained model in an inertial localization scheme.

Previous works on DR  techniques emphasize the importance of accurate speed updates to reduce the positioning error. Brossard et al. proposed special speed updates when specific events are detected using machine learning or other methods \cite{brossard2019rinsw}. Key example is the zero velocity update (ZUPT) when a full stop is detected. Other examples include nullifying the vertical and horizontal components of the velocity of a four wheeled vehicle when no slip is detected. The detection of special events with respective corrections to the velocity vector are used to nullify accumulated integration errors. The approach proposed in this work is a generalization of this idea because the trained model can learn all possible detectable events with their respective speed updates to automatically reset accumulated errors.

The rest of the paper is organized as follows: Section II defines the DR problem with speed as input, Section III presents the learning approach, including a highly accurate data collection method and DNN architecture. Section IV presents the model training results with hold-out set evaluation and application as input for DR localization, Section V gives the conclusions.





\section{Dead Reckoning with Speed Input}



For testing the proposed approach, a two-dimensional DR scheme is used. The speed and the heading angle of the car (both with respect to navigation frame) are used in a single integration step to obtain the position. Consider the following two-dimensional linear kinematic equation of motion for the car position
\begin{equation}
\left( {\begin{array}{*{20}{c}}
{p_x^n}\\
{p_y^n}
\end{array}} \right)\left[ k \right] = \left( {\begin{array}{*{20}{c}}
{p_x^n}\\
{p_y^n}
\end{array}} \right)\left[ {k - 1} \right] + \Delta t\left( {\begin{array}{*{20}{c}}
{v_x^n}\\
{v_y^n}
\end{array}} \right)\left[ k \right]
\label{eq:pos}
\end{equation}
where $p_i^n\left[ k \right]$ is the car position and $v_i^n\left[ k \right]$ is the car velocity in the $i \in \left\{ {x,y} \right\}$ axis, at time step $k$. The $n$ denotes the navigation coordinate frame and $\Delta t$ is the time step size. The car velocity is provided by
\begin{equation}
\left( {\begin{array}{*{20}{c}}
{v_x^n}\\
{v_y^n}
\end{array}} \right)\left[ k \right] = \left( {\begin{array}{*{20}{c}}
{\cos \psi }\\
{\sin \psi }
\end{array}} \right)\left[ k \right] s^b\left[ k \right]
\label{eq:vel}
\end{equation}
where $\psi\left[ k \right]$ is the car heading  angle and $s^b \left[ k \right]$ is the car speed at time step $k$ in the body frame. The car heading angle $\psi\left[ k \right]$ is obtained using an attitude estimation algorithm \cite{farrell2008aided} which also provides the roll and pitch angles used to transform IMU measurements from the body to navigation frame. The car speed $ s^b\left[ k \right]$ can be obtained in two ways: (1) from the output of the trained DNN model and (2) by direct integration of the noisy acceleration signal (in navigation frame).

 
\section{Learning Approach}
\subsection{Motivation}
It is widely known that the direct integration of acceleration measured with a low-cost IMU sensor to obtain speed (or velocity) results in rapid divergence due to measurement noise \cite{farrell2008aided}. This section describes the development of a DNN-based car speed estimator with noisy IMU measurements as input. A supervised learning approach is presented where the IMU measurements are labeled with highly accurate speed measurement, using the RTK-GNSS device,  which the DNN model is trained to predict.


\subsection{Dataset collection and generation}
\begin{figure}[h!]
\centering
{\includegraphics[width=0.4\textwidth]{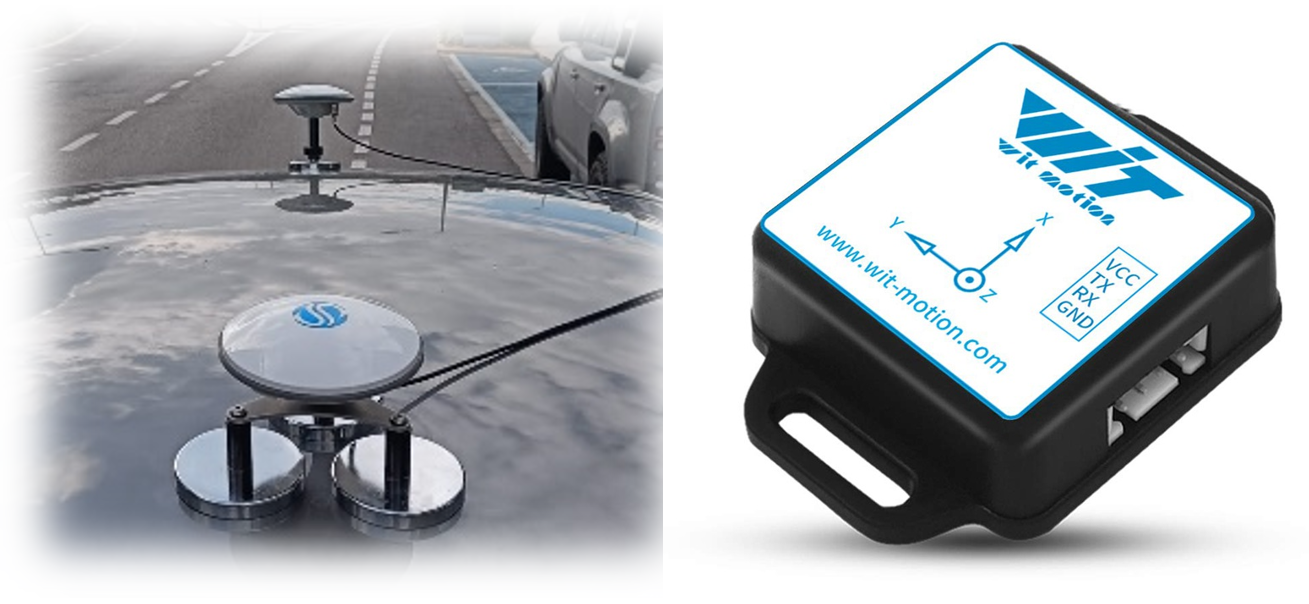}}
\caption{STONEX RTK-GNSS device (left) and WitMotion BWT61CL IMU sensor (right).}
\label{fig:RTKIMU}
\end{figure}

\begin{figure}[h!]
\centering
{\includegraphics[width=0.35\textwidth]{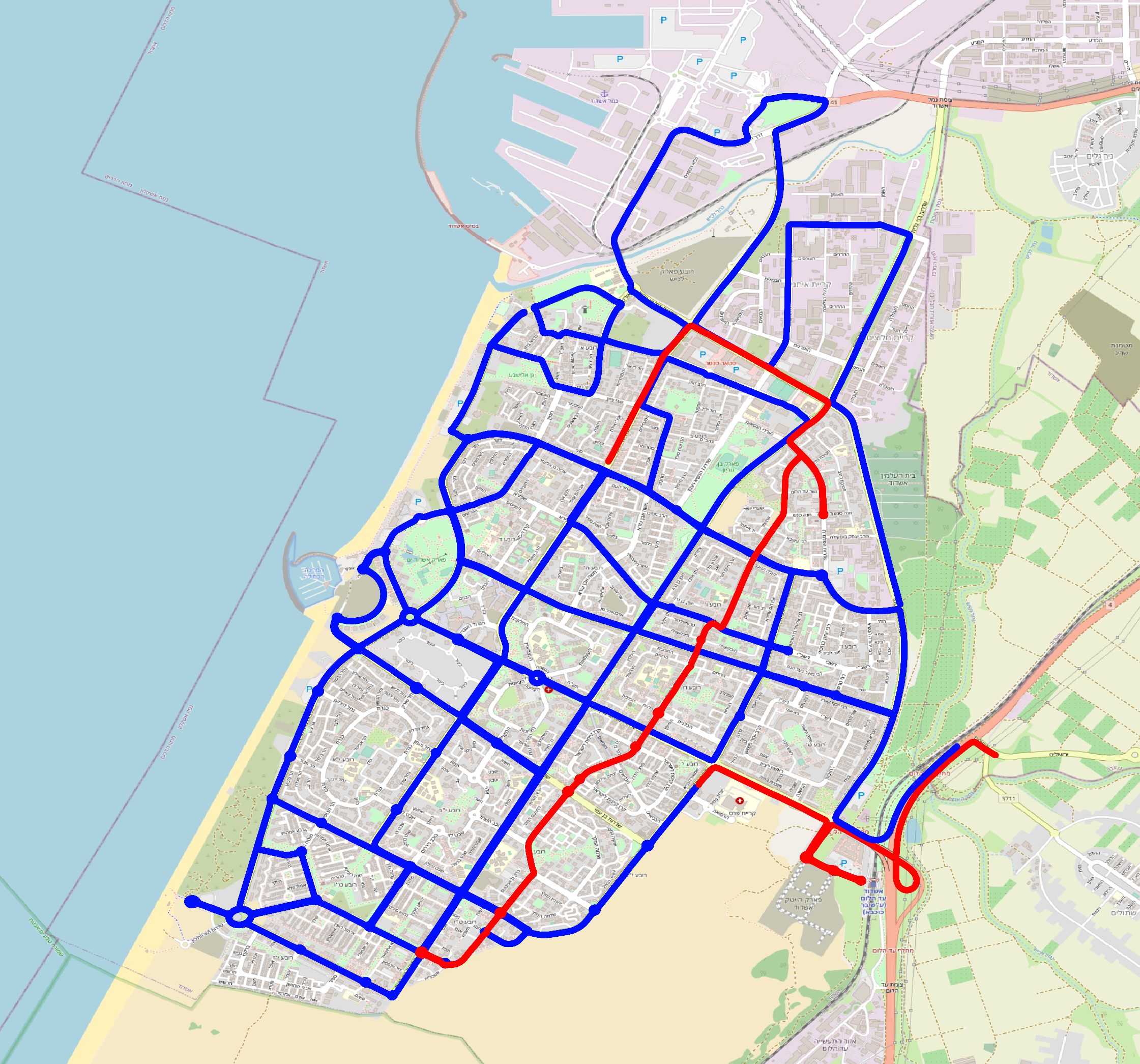}}
\caption{Car trajectory for the train and validation sets in an urban area (Ashdod, Israel). No overlap between the train and validation data sets.}
\label{fig:datacollection}
\end{figure}

Data was collected with a WitMotion BWT61CL IMU sensor \cite{wit_imu} and a STONEX  RTK-GNSS device connected to a Windows PC through serial port connections (Figure \ref{fig:RTKIMU}). Data was recorded at $50 \, Hz$ and $100 \, Hz$ rates with the RTK-GNSS and IMU sensors, respectively. The sensors were synchronized to a single clock. Accurate ground truth labels are critical in supervised learning tasks. The RTK-GNSS STONEX device provides position measurements with a $2 \, cm$ average positioning error in contrast to the common GNSS receiver which has an average error on the order of meters \cite{farrell2008aided}. A total of $180$ minutes of driving was recorded in Ashdod, Israel. The recordings were split into train and validation sets with a ratio of 85:15. Figure \ref{fig:datacollection} shows the trajectories of the recorded drives for the train and validation sets (without overlap). The data includes diverse driving with numerous turns and roundabouts at speeds typical to city driving. 

\begin{figure}
\centering
{\includegraphics[width=0.48\textwidth]{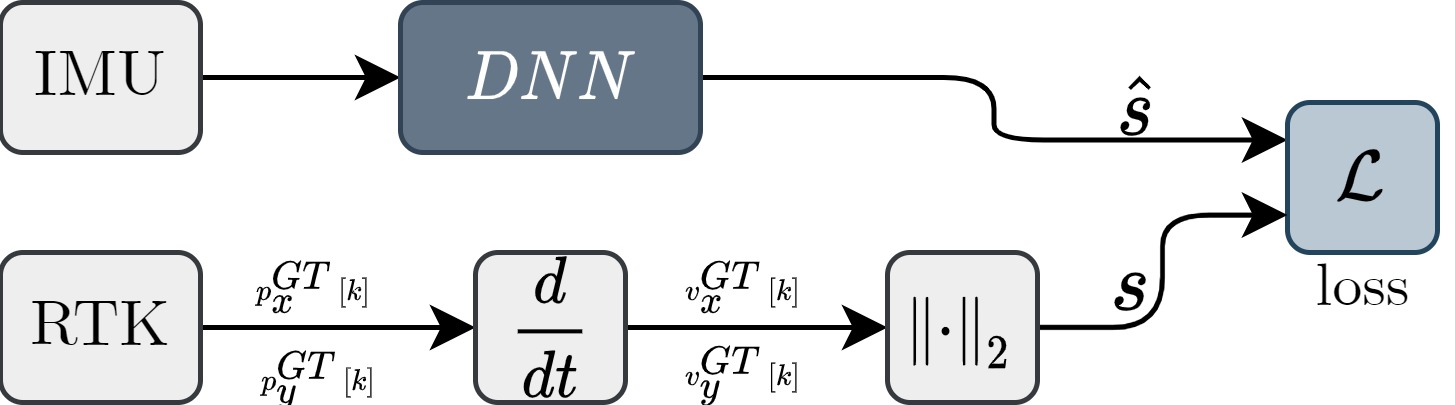}}
\caption{Car speed regressor training diagram. The IMU and GNSS-RTK signals are input to the loss function where the first  goes trough the DNN and the later used to derive the car speed.}
\label{fig:training_diagram}
\end{figure}

Speed was obtained from the position measurements by taking the derivative with respect to time for each component and then calculating the vector norm as shown in Figure \ref{fig:training_diagram}. Then, the 50 Hz speed signal was upsampled to match the 100 Hz rate of the IMU. Train and validation samples were created by matching windows of 20 measurements of the 6 channel IMU with their respective window of 20 samples of speed. Acceleration measurements were rotated to navigation frame to remove the gravity component and then rotated back to body frame.



\subsection{LSTM based-model architecture}
\begin{figure}[h!]
\centering
{\includegraphics[width=0.5\textwidth]{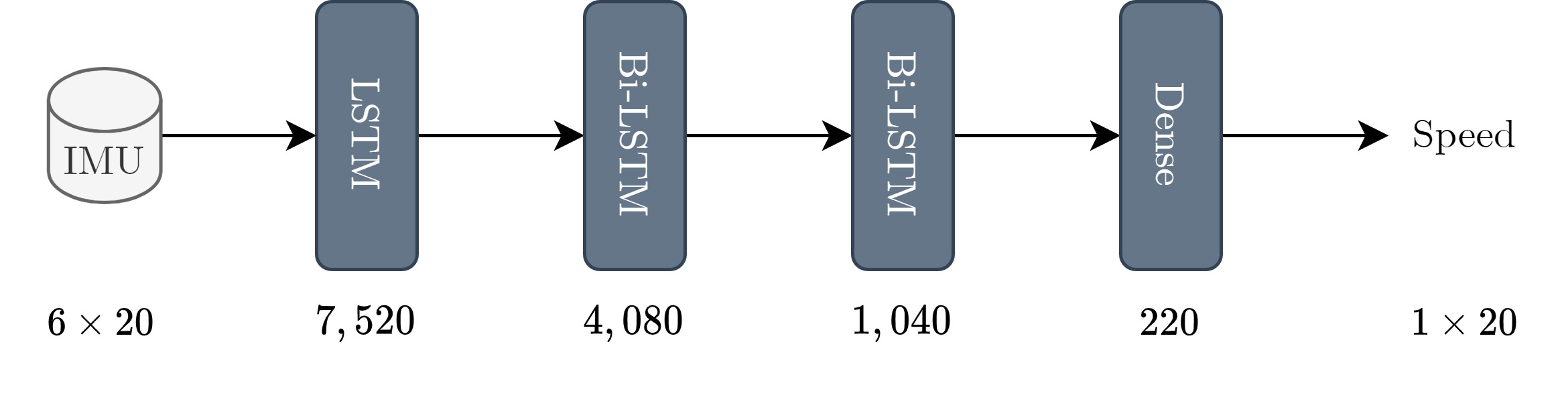}}
\caption{DNN architecture with many-to-many input-output relation, receives 20 samples of a 6-channel IMU as input and returns 20 samples of speed prediction.}
\label{fig:fig_dnn}
\end{figure}

A DNN architecture was selected comprising one LSTM layer, two bi-directional LSTM layers, and a single dense layer. The model contains a total of 12,860 trainable parameters. The model is of type "many-to-many", i.e., each prediction expects 20 samples of a 6 channel IMU as input and returns 20 samples of speed prediction. The architecture and input dimension are summarized in Figure \ref{fig:fig_dnn}. The LSTM layers were configured to pass the hidden state between samples to allow long term memory learning between predictions.



\subsection{Loss function and training}
The following mean square error (MSE) loss function was minimized in the training stage
\begin{equation}
{\cal L} = \frac{1}{{2N}}\sum\limits_{j = 1}^N {{{\left( {{\bf s}_j^{GT} - {{{\bf \hat s}}_j}\left( {{\bf{a}},{\bf{\omega }};{\bf{W}}} \right)} \right)}^2}}
\label{eq:loss}
\end{equation}
Here, $\bf \hat s$ is a function of ${\bf{a}}$, ${\bf{\omega}}$, and the network weights $\bf W$. ${\bf{a}}$ and ${\bf{\omega}}$ are the accelerometer and gyroscope batch readings. The training data was divided into 4 drives each about 40 minutes long. Each drive was broken into batches ordered by acquisition time to allow the model learn long term memory. Each training batch consisted of 4 windows of $6 \times 20$ IMU measurements (with the gravity component subtracted) that followed the previous batch and passed forward the LSTM hidden states. Shorter drives were padded with zeros (which the loss function was modified to ignore) to match the drive time. At the start of each epoch the LSTM hidden states were reset to zeros. 


\section{Results}

\subsection{Training results}
\begin{figure}[h!]
\hspace*{-1cm}
\centering
\includegraphics[scale=.21]{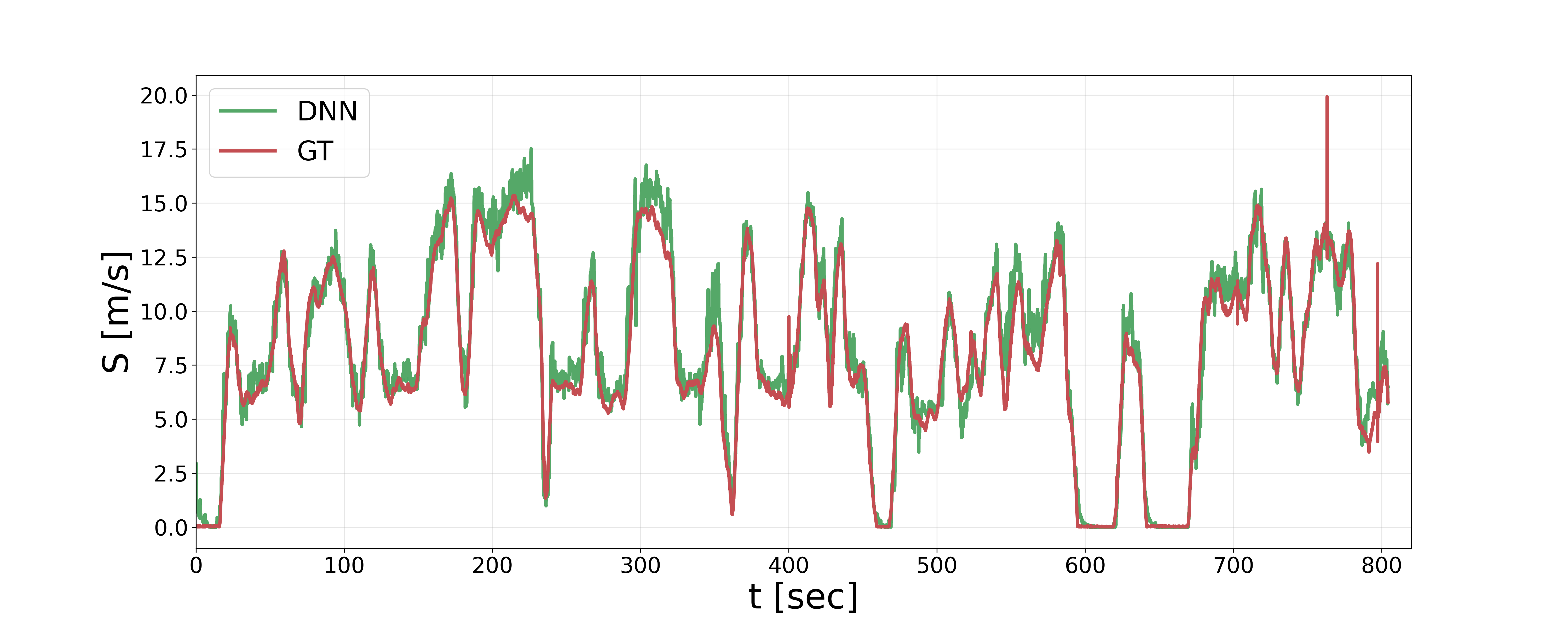}
\caption{Estimated (DNN model) and true car speed vs. time for a single drive from the validation set (RMSE = $0.90 \, m/s$).}
\label{fig:fig_speed_val}
\end{figure}

The model was trained until convergence for 200 epochs. The root MSE (RMSE) reached a value of $0.83 \, m/s$ on the train set and $1.21 \, m/s$ on the validation set. Figure \ref{fig:fig_speed_val} shows a representative example of speed prediction from the validation set. The above RMSE on the validation set may be interpreted as a 12 meter positioning error during a $10 \, s$ drive. For a drive at an average speed of $15 \, m/s$, the car passes a total $150 \pm 12 \, m $. In this case the positioning error is $8\%$ of the driven distance.

\subsection{Model application in aided DR navigation}
The trained DNN model is tested on a two-dimensional navigation problem using a hold-out dataset. A 4 minute drive in the Matam high-tech park, Haifa was recorded using the same equipment and car. The train/validation and the hold-out sets were recorded in relatively flat urban areas. Position is obtained by integrating the car speed after decomposition to velocity vector using the heading angle as shown in Equations \ref{eq:pos} and \ref{eq:vel}. For the integration, speed was obtained in two ways: (1) using the trained model and (2) by direct integration of the acceleration in navigation frame (used as reference).

\begin{figure}[h!]
\centering
{\includegraphics[width=0.5\textwidth]{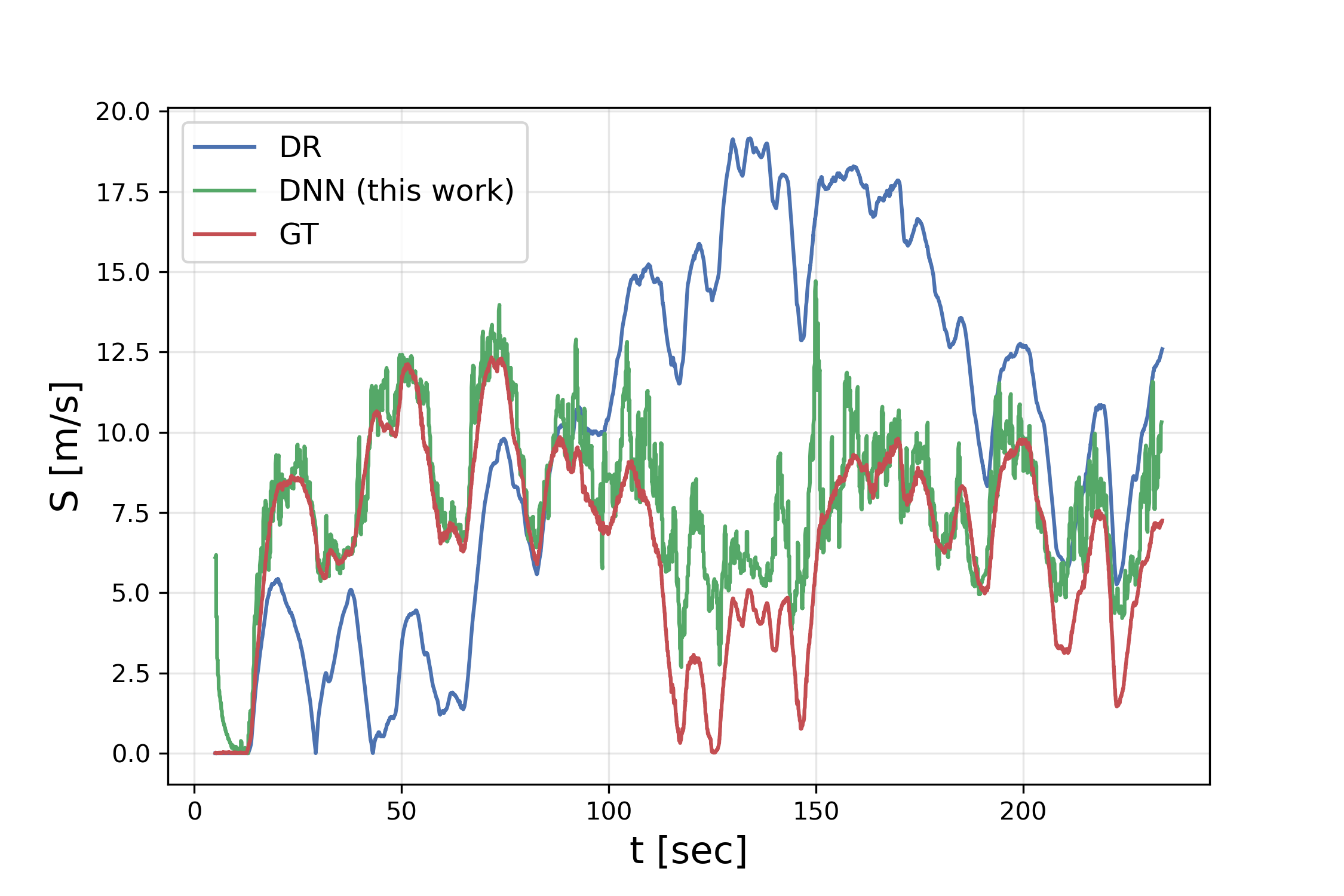}}
\caption{Car speed vs. time (test set): dead reckoning (DR), trained DNN model, and ground truth (GT).}
\label{fig:speed_comparison}
\end{figure}

Figure \ref{fig:speed_comparison} shows the speed of the car versus time obtained by direct integration of acceleration (in navigation frame), DNN prediction, and the ground truth (RTK-GNSS). The car starts stationary and the DNN prediction takes several seconds to nullify despite starting with hidden states set to zero. The first acceleration starts near $t = 10 \, s$ and the DR speed diverges rapidly from the ground truth. The DNN speed follows the ground truth with good agreement until $t = 110 \, s$ (RMSE $(0s<t<110s)= 1.08 \, m/s$). At that point the road quality was observed to be low and significant external vibrations were introduced to the signal (RMSE $(110s<t<150s) = 3.30 \, m/s$). At $t = 150 \, s$ the car finished a roundabout and returned to the high quality asphalt which resulted in better speed prediction accuracy (RMSE $(150s<t<233s)= 1.72 \, m/s$). 



\begin{figure}[htp]
\centering
{\includegraphics[width=0.5\textwidth]{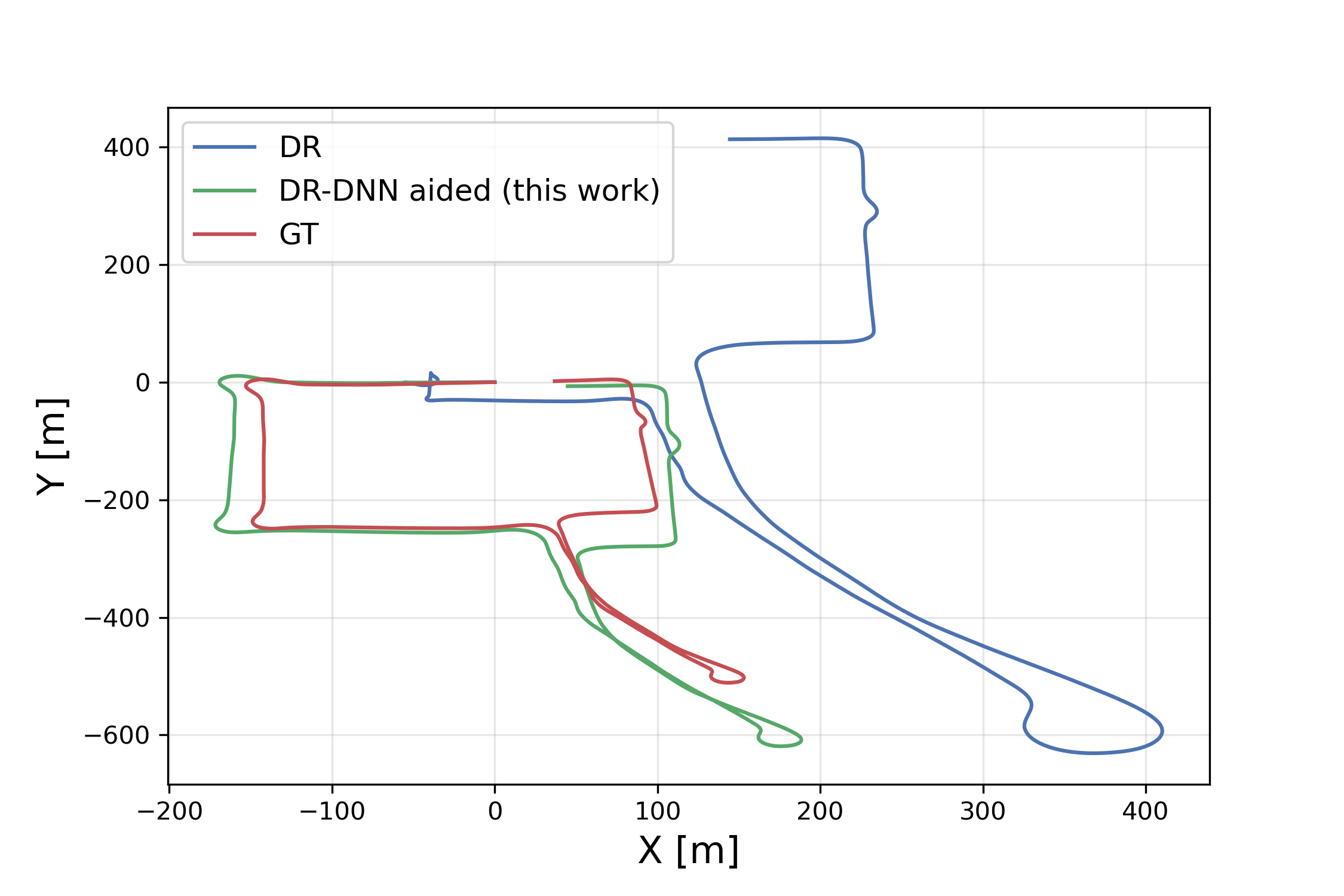}}
\caption{Car trajectory calculated using plain and aided DR vs. ground truth.}
\label{fig:fig_trajectory}
\end{figure}

\begin{figure}[h!]
\centering
{\includegraphics[width=0.5\textwidth]{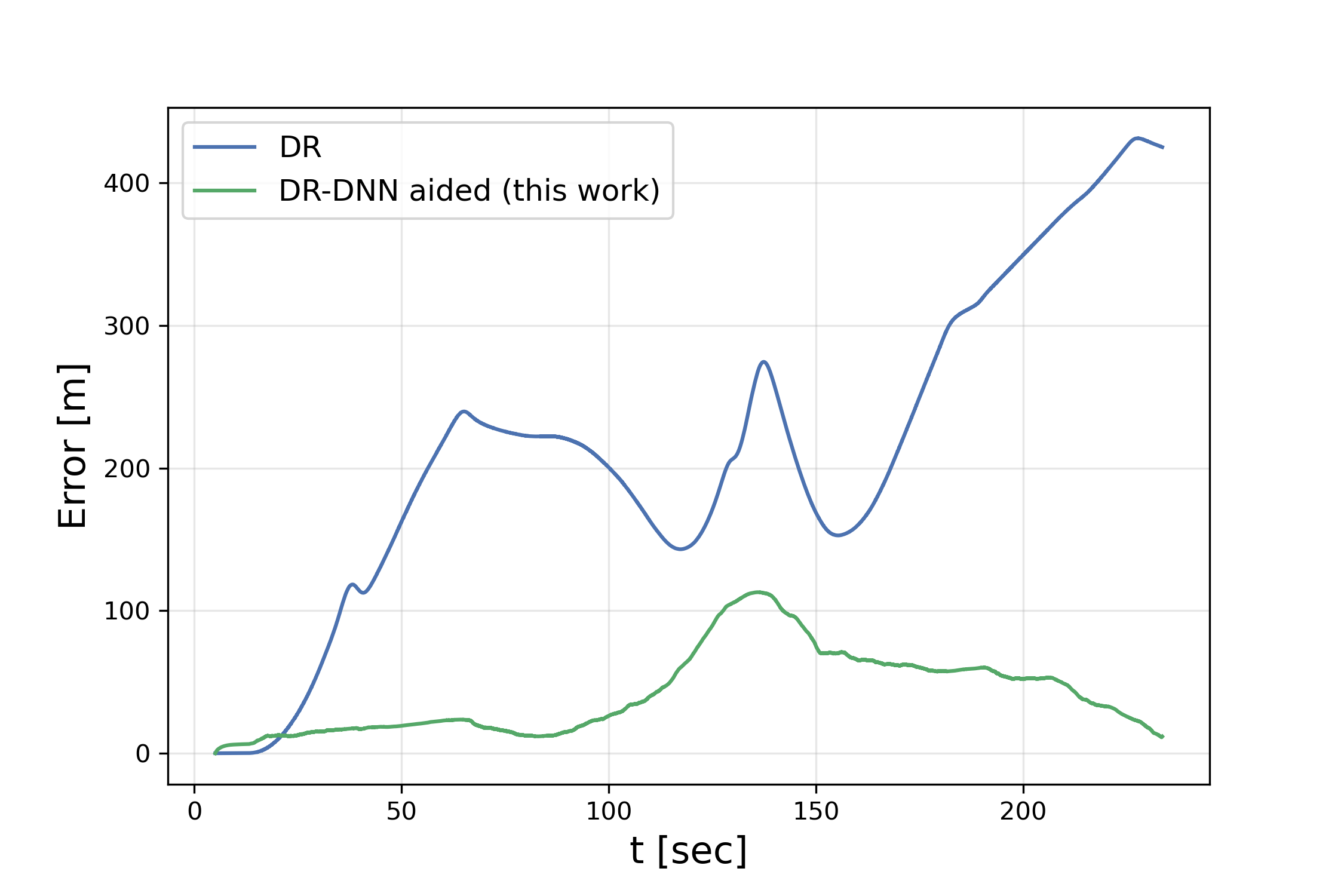}}
\caption{Position error vs. time using plain and aided DR.}
\label{fig:fig_error}
\end{figure}

Figure \ref{fig:fig_trajectory} shows the car trajectory obtained using the two different methods and the ground truth. The plain (without DNN aid) DR fails to follow the car true trajectory at the drive start, which was expected due to the speed error shown in Figure \ref{fig:speed_comparison}. At later time steps the shape of the DR curve becomes more like the ground truth. However, this is mainly because of the heading angle accuracy which defines the shape but not the length of the segments. The aided DR (DL-DNN aided) shows good agreement with the ground truth and a substantial improvement over plain DR. The significant deviation from true speed is observed near the roundabout located at the southern point of the trajectory. Figure \ref{fig:fig_error} shows the error versus time for the two position solutions. The error is defined as the distance between the position solution and the ground truth. The plain DR diverges rapidly over time while the DNN aided scheme maintains less than $20$ m error for more than a minute and less than $120$ m during the whole 4 minute drive.

\section{Conclusions}
A deep neural network model was trained to estimate the speed of a car from noisy IMU measurements. A new method for collecting accurate speed measurements at high frequency was presented using an RTK-GNSS device. The trained model is light (less than 13,000 trainable parameters) and suitable for real-time applications. A simplified aided dead reckoning localization scheme was used to asses the performance of the trained model. The DNN provided speed pseudo-measurements which substantially improved the positioning accuracy over the direct integration approach. The proposed methodology can be adapted and extended to other ground vehicle and pedestrians as well as aerial and water vehicles. In future work, more data will be collected to introduce additional driving scenarios, rich terrain properties, and non-flat roads. Additionally, real-time implementation for GNSS-free navigation is currently under development. 

\normalsize

%
%

%
%

\bibliographystyle{IEEEtran}
\bibliography{IEEE_bib}

\end{document}